\documentclass[10pt,twocolumn,letterpaper]{article}

\usepackage{cvpr}              
\usepackage{algorithm}
\usepackage{algpseudocode}

\definecolor{cvprblue}{rgb}{0.21,0.49,0.74}
\usepackage[pagebackref,breaklinks,colorlinks,allcolors=cvprblue]{hyperref}

\def\paperID{17539} 
\def\confName{CVPR}
\def\confYear{2025}

\title{From Videos to Indexed Knowledge Graphs - Framework to Marry Methods for Multimodal Content Analysis and Understanding}

\author{Basem Rizk\\
University Of Southern California\\
Los Angeles, CA\\
{\tt\small basem.rizk@outlook.com}
\and
Joel Walsh\\
University Of Southern California\\
Los Angeles, CA\\
{\tt\small jwalsh@ict.usc.edu}
\and
Mark Core\\
University Of Southern California\\
Los Angeles, CA\\
{\tt\small core@ict.usc.edu}
\and
Benjamin Nye\\
University Of Southern California\\
Los Angeles, CA\\
{\tt\small nye@ict.usc.edu}
}

\begin{document}
\maketitle
\begin{abstract}
Analysis of multi-modal content can be tricky, computationally expensive, and require a significant amount of engineering efforts. Lots of work with pre-trained models on static data is out there, yet fusing these opensource models and methods with complex data such as videos is relatively challenging. In this paper, we present a framework that enables efficiently prototyping pipelines for multi-modal content analysis. We craft a candidate recipe for a pipeline, marrying a set of pre-trained models, to convert videos into a temporal semi-structured data format.  We translate this structure further to a frame-level indexed knowledge graph representation that is query-able and supports continual learning, enabling the dynamic incorporation of new domain-specific knowledge through an interactive medium.
\end{abstract}    
\section{Introduction}
\label{sec:intro}
One way of analyzing multi-modal content is integrating the diverse interpretation of available information modalities. For instance with videos, these correspond to the visual, auditory, and textual channels. Transforming any multi-modal and particularly temporal complex data into a structured \cite{DBLP:journals/corr/abs-2007-10040}, query-able format remains an open problem.

Existing neural methods focusing on particular use cases, with help of the vast amount of data available nowadays, have gotten pretty good at analyzing individual modalities. However there is a need for a more comprehensive approach that can effectively combine these modalities and represent them in a structured manner, with temporal understanding \cite{PEI20131369}, in a fashion similar to manually designed expert systems \cite{5206492}. This conceptually brings us back to the earlier days of AI, along with the pursuit of explainable expert systems. Early Expert systems aimed to gather and organize human expertise in a way that computers could utilize. Although they might work on a small scale, they were extremely expensive to build, and they often struggled with the complexity and tractability of real-world problems. More importantly, we need to have the capability of continual learning to adapt to domain-specific knowledge' shifts and specifics.    

This paper presents a novel approach to constructing and querying knowledge graphs from video data, with a focus on continual learning and knowledge extension. We introduce a framework that facilitates the integration of various pre-trained models and methods for multi-modal content analysis. Our main contributions \footnote{Our implementation is available at \url{https://github.com/ICTLearningSciences/content-analysis-playground}} in this paper to the domain of analysis, understanding, and searching in videos consist of the following:
\begin{itemize}
    \item We designed and implemented a framework to build pipelines to marry methods and inferences for multimodal content understanding and analysis.
    \item We designed a novel method and implemented a proof-of-concept software to query information from a database of videos while being able to add new domain-specific knowledge/annotation to it too. This method employs a candidate pipeline recipe that we crafted and built using our framework, consisting of a combination of pre-trained models and existing methods.
\end{itemize}

\section{Related Work}
Great efforts has been put into image captioning \cite{li2022blipbootstrappinglanguageimagepretraining}. These efforts has led to even question-answering capabilities over images \cite{li2023blip2bootstrappinglanguageimagepretraining}. Moreover, with the promising and continuously intriguing performance of Large Language Models (LLMs), more work has recently started going into visual-grounding and modeling, whether by fusing the learned space of each modalities \cite{kim2024openvlaopensourcevisionlanguageactionmodel, meta_ai_llama3}, or by having a shared embedding space across all modalities \cite{openai_gpt4o}.

Knowledge graph extraction from text has traditionally involved a pipeline of smaller tasks, starting with term extraction. Term extraction methods help identify knowledge graph entities  through statistical \cite{5392697}
, information theoretic \cite{inproceedings}
, or neural unsupervised learning \cite{CAMPOS2020257} . Relation Extraction methods take pairs of these raw entities and determines if the sentence or document implies that a relation exists between the two entities. This is often framed as a neural classification task using semantic features \cite{yao2019docred} \cite{zhang2017tacred}. 
Recently there has been some success  using LLMs to complete the entire task of extraction  from text end-to-end \cite{edge2024localglobalgraphrag}. Previous attempts at extracting knowledge graphs from videos have used annotations as a source text \cite{mahon2020knowledgegraphextractionvideos}.

In pipelines, utilizing generated data for building datasets, there have been challenges with respect to noise inherited by the extracted relationships and elements. On the work of conditional image similarity \cite{vaze2023genecisbenchmarkgeneralconditional}, a methodology was introduced for extracting relationships from images by using a scene-graph parser  \cite{scene_graph_parser, schuster-etal-2015-generating}  to identify subject-predicate-object triplets within image captions. These triplets are then filtered using concreteness scores \cite{Brysbaert2014ConcretenessRF}, which measure how tangible or concrete the subject and object are.  This filtering process helps to remove noisy or irrelevant relationships, improving the quality of the extracted information.  

Caption Anything \cite{wang2023caption} is an interactive system to caption target details of an image on command. This system employs the state-of-the-art segmentation model, Segment Anything (SAM) \cite{kirillov2023segany} --- a segmentation model that accepts prompts including but not limited to a bounding box, or a point in the space of the image, to detect the user targeted segment of an image. They use the mask generated by SAM to create an image with only the masked area appearing, while the rest of the image is blank. This partially-blinded version of the image is then passed to a captioner module employing BLIP2 \cite{li2023blip2bootstrappinglanguageimagepretraining}.
 
Since the development of AlexNet \cite{NIPS2012_c399862d} on ImageNet \cite{imagenet_cvpr09}, researchers have been able to build high performing classifications models to differentiate and detect classes of objects. Recently, along with the popularity of vision language models (VLMs), and incorporating backbones of self-supervised cross-modality embedding models, more work has gone into open-set detection models. GroundingDino \cite{liu2023grounding} incorporates self-supervised vision transformers \cite{DBLP:journals/corr/abs-2104-14294}, and accepts textual prompts consisting of open-vocabulary categories, or referring expressions (e.g., the person sitting on a chair). Furthermore, another recent model is Recognize Anything (RAM) \cite{zhang2023recognizeanythingstrongimage} which has shown impressive performance in image tagging. The potential of RAM have caught researchers attention to construct prompts for GoundingDino for a complete objects detection and localization pipeline \cite{ren2024groundedsamassemblingopenworld}.

Visual ChatGPT \cite{wu2023visualchatgpttalkingdrawing} have also combined text-only LLM capabilities of ChatGPT with vision foundational models (VFMs) as BLIP \cite{li2023blip2bootstrappinglanguageimagepretraining} and Pix2Pix, using a Prompt Manager that serves somewhat as an interface of the visual channel of the input. It requires heavy amount of prompt engineering to translate each foundational model outcomes into tangible value in textual format. Another similar work is DetGPT \cite{pi2023detgpt} by incorporating zero-shot performing frozen model, GroundingDino.
\section{Methodology}
Our methodology consists of 3 phases. First, we built a framework that allows running optimized compositions of pre-trained models to experiment quickly and plug and play with open-source models to process temporal multi-modal data such as videos. Second, utilizing our framework, we designed a pipeline to transform videos into a semi-structured data format, `\textbf{VideoKnowledgeBase}'(s), employing qualitatively selected pre-trained models and existing methods. Third, we designed an algorithm to transform the generated VideoKnowledgeBases into Video Knowledge Graphs, which are query-able, and extensible using mini-classifiers. In this section, we describe each of the three phases in details.

\subsection{Framework to Marry Inferences \& Methods}\label{subsec:method_framework}
Our aim is to build a framework that allows pipelines to incorporate and marry various methods and/or models seamlessly. These pipelines shall be capable to process temporal data in at least near real-time, under the assumption that it is theoretically feasible to do so, and practically each of those methods and models can independently run in real-time. The framework must be flexible enough to allow the flow of any form of inferences from a stage (`\textbf{Pipe}') to another in the pipeline. 

\begin{figure*}
    \centering
    \includegraphics[width=0.95\linewidth]{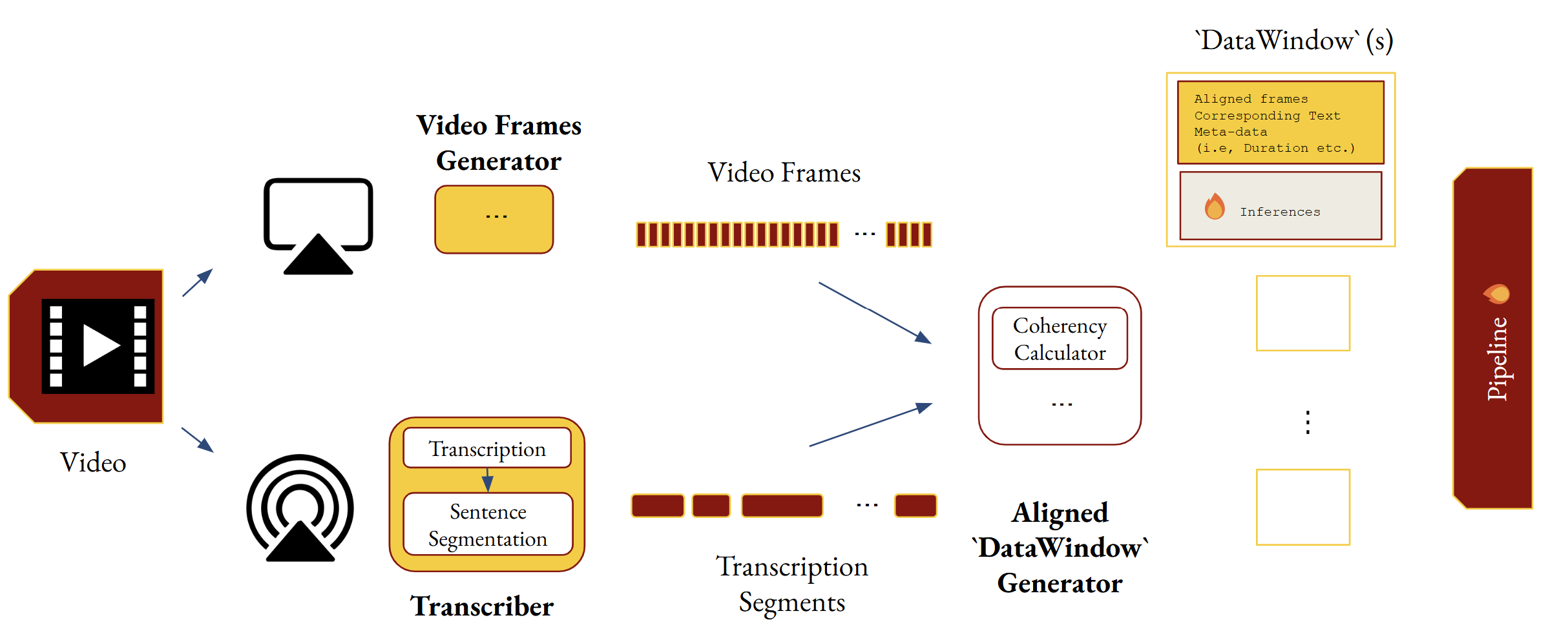}
    \caption{Pre-pipeline: An illustration of an example of a `\textbf{DataWindowGenerator}`. This DataWindowGenerator in the figure particularly accepts a video, transcribes it, and segments the video on the basis of the transcription paragraphs. Those paragraphs are constructed utilizing a greedy approach using coherency scores. It yields DataWindows that packs aligned segments of frames's images with corresponding segments of coherent segments of transcription.}
    \label{fig:datawindow_gen_prepipeline}
\end{figure*}

Hence we define a `\textbf{DataWindow}', which serves as a logical unit to encapsulate a segment of multi-media. Typically there will be a DataWindow generator that would generate the earliest form of those DataWindow instances. To process a video for example, the DataWindow generator might work as illustrated in figure \ref{fig:datawindow_gen_prepipeline}, where the DataWindow corresponds directly to a segment of frames from the video across time, which is aligned with a textual segment of transcription. DataWindow instances flow from one Pipe to another. A DataWindow contains theoretically an unlimited number of placeholders for any format and any type of inferences, allowing Pipe(s) to read, and add/manipulate information (e.g., model inferences) as needed by the business logic.

As mentioned above, a Pipe is a processing component that takes a DataWindow as input and yields/returns a DataWindow as output. It would be used typically to wrap a machine learning model for inference so that it can team up with other Pipes in a pipeline (e.g., Use some other model(s) outcome/inference as input, which will be found in the DataWindow).
For convenience and scalability, we further define a `\textbf{PipeDirector}' --- in analogy to a movie director directing the way a scene is being acted out, it directs the application of the pipe method onto the DataWindow. The PipeDirector is responsible to extract the data to be lightly preprocessed and reformatted, typically based on common interfaces (i.e., bu in PipeDirectors logic) to match the employing Pipe's input format. While the PipeDirector is an implementable interface, we have built few types of PipeDirectors, which are listed in our supplementary material.

We designed the Pipes to work in a micro-service fashion. Moreover, to allow seamless integration and to utilize managing the flow of the DataWindow(s) across those pipes, we define an orchestrator component, `\textbf{Pipeline}' with its variations: `Sequential', `Parallel', and `Loop'. Sequential and Parallel pipelines shall run as their names denote, while the Loop pipeline would be meant to repass the DataWindow across a sequence of Pipes until a condition is met. Along with the purpose of each of those variation, a main benefit in creating a Pipeline acting as a manager for a sequence or a group of Pipes is ``pipeline parallelism,'' where we abstract the processing logic to maximize utility of resources to attempt to achieve near real-time performance.

Furthermore, for more layers of abstraction, as we did by creating variations of typical variations of PipeDirectors, we did the same with Pipes by defining variations. To draw a picture, some of which are: `BatchInferencePipe,' that utilizes employing batch-compatible inferences methods, `BranchingInferencePipe,' that maps each single input into many outputs (i.e., an image to many augmented versions of this image), and `MergingInferencePipe' to handle the opposite of the former.

Given that Pipes accept only DataWindow(s), a component that acts as an input adapter of any Pipeline is `\textbf{DataWindowGenerator}'. It can be observed as semi-pipe, as it outputs DataWindows; however it accepts whatever type of data it is designed to deal with. In our use-case, we have defined an example of DataWindowGenerator, illustrated in figure \ref{fig:datawindow_gen_prepipeline}, that accepts a video, transcribes it and segments it on the basis of the transcription. Moreover, it packs those segments of video while aligned with the transcription into DataWindows in sequentially in run-time. It is explained in more details in the following subsection \ref{subsec:method_recipe}. A Pipeline accepts accordingly a DataWindow or a DataWindowGenerator.

\begin{figure*}
    \centering
    \includegraphics[width=0.95\linewidth]{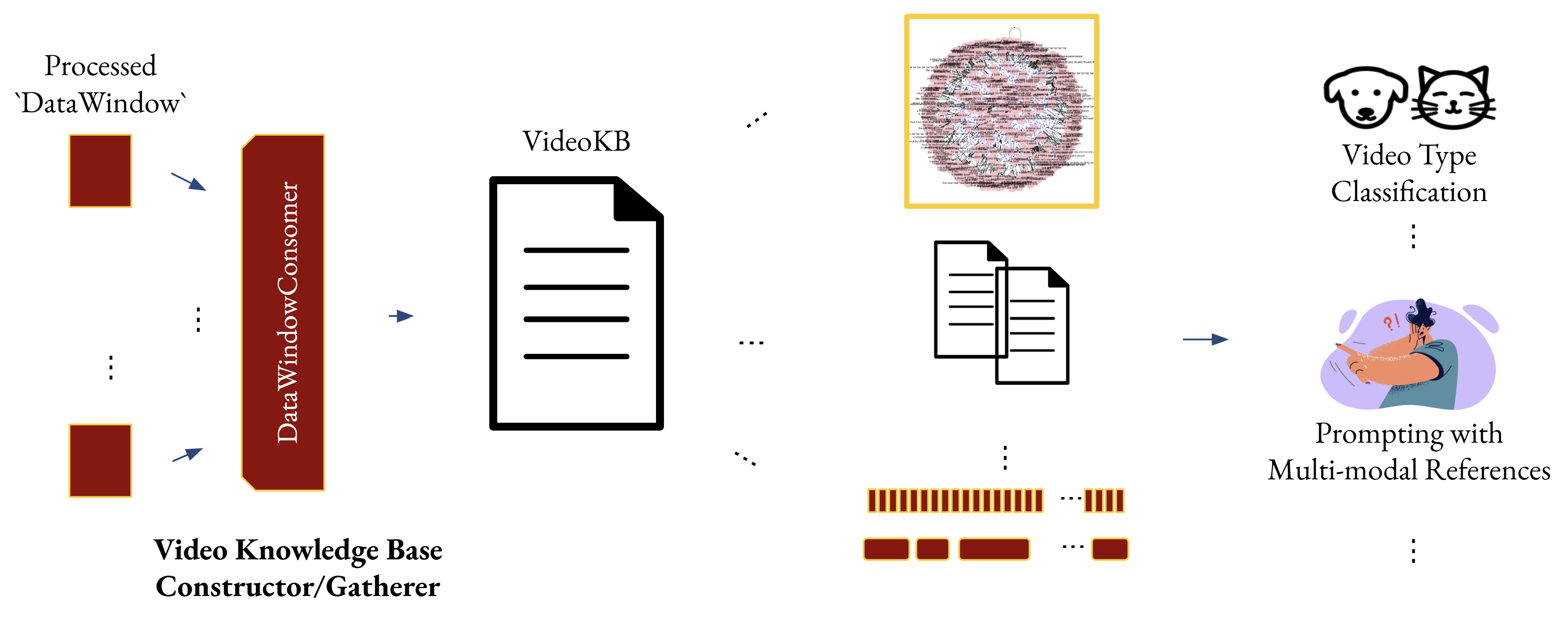}
    \caption{Post-pipeline: Abstract illustration of how a `\textbf{DataWindowConsumer}` writing the DataWindows of a video into a semi-structured format which we call VideoKnowledgeBase. That is to be utilized by downstream tasks (e.g., video type classification, information retrieval, generating knowledge graphs). 
    }
    \label{fig:datawindow_consume_postpipeline}
\end{figure*}

At the tail of the Pipeline, as it is outputting DataWindows, we define the concept of `\textbf{DataWindowConsumer},' illustrated in figure \ref{fig:datawindow_consume_postpipeline}. It could well be perceived as a semi-pipe, as it accepts DataWindow(s), and outputs or writes any data format needed for the business logic. For our use case, and to utilize employed models' inferences for further analysis, we write video sequence DataWindows in a format that we refer to as VideoKnowledgeBase, which can be used then for experimentation on utilizing this information in downstream tasks (e.g., video type classification, information retrieval, generating knowledge graphs).




\subsection{Recipe: Videos into Semi-structured Data} \label{subsec:method_recipe}
\begin{figure*}
    \centering
    \includegraphics[width=0.8\linewidth]{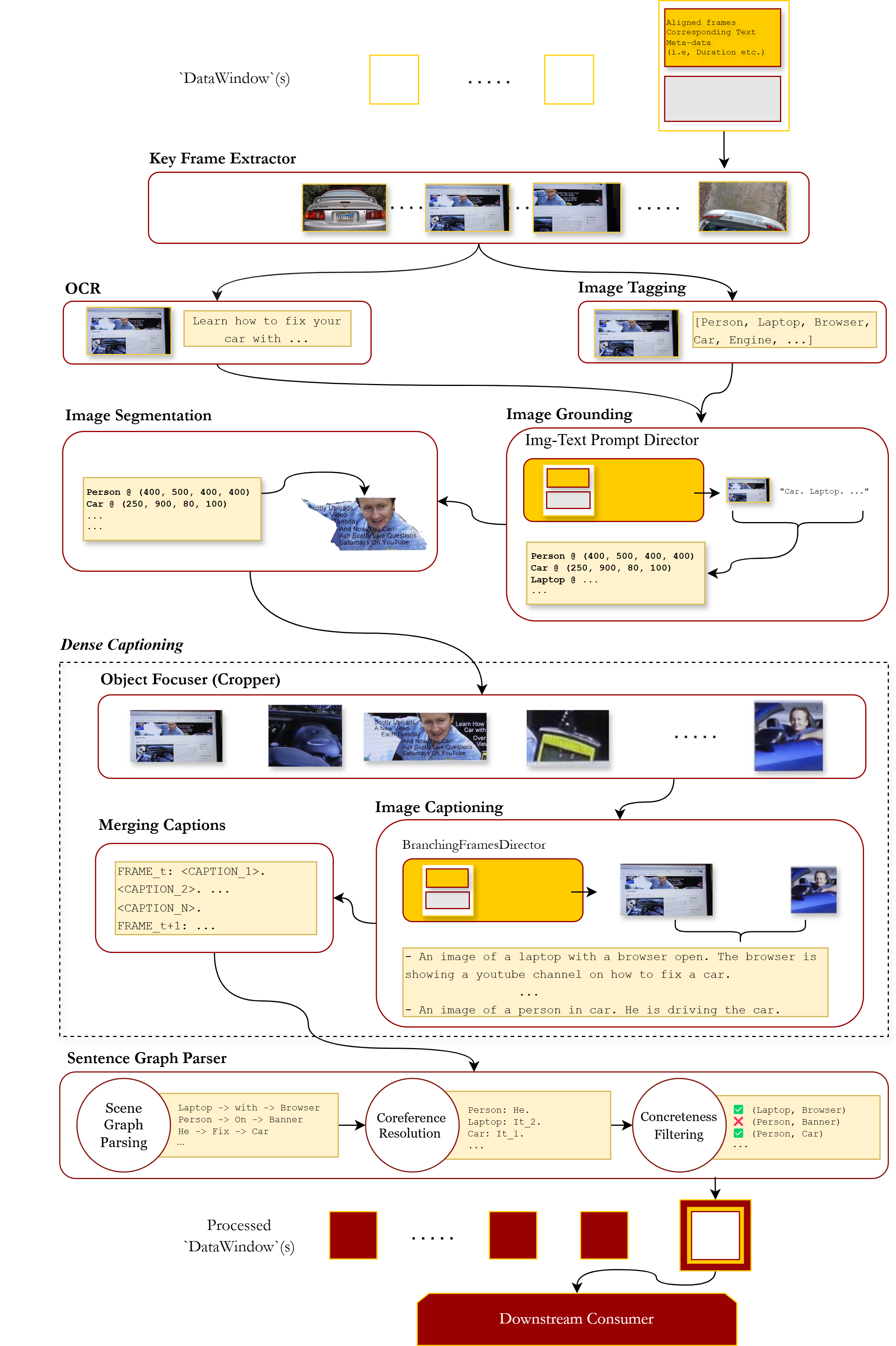}
    \caption{An illustration of our pipeline recipe and what we aim to achieve by each of employing a combination of pre-trained models. The pipeline transforms video data into a semi-structured knowledge base, begining with extracting keyframes from the video and then applying various computer vision techniques, such as OCR, image tagging, and dense captioning. The resulting information is then processed to extract relationships between objects and entities, which are used to construct a knowledge graph.
    }
    \label{fig:pipeline}
\end{figure*}

Employing our implemented framework per the design explained in subsection \ref{subsec:method_framework}, we craft a pipeline recipe, illustrated in figure \ref{fig:pipeline}, to process videos and transform video data into semi-structured data, which allows further analysis and content understanding research.

\subsubsection{Generating DataWindows}
As explained in the previous section, our Pipeline framework understands DataWindows. For our use case, we built a DataWindowGenerator for videos, illustrated in figure \ref{fig:datawindow_gen_prepipeline}. Our generator takes in a video and generates sequentially DataWindows, each packing ordered coherent segments of the video. We get timestamped transcription of the video, using OpenAI's Whisper model \cite{radford2022whisper}. We intentionally avoid relying on Whisper segmentation, as it does not segment many of the sentences. Instead of we apply explicitly sentence segmentation over the whole transcription employing a Spacy pipeline \cite{spacy_honnibal2020} incorporating "en\_core\_web\_lg" language model, and then we realign the word-level timestamps from the transcription. Furthermore, we utilize coherency scoring model by calculating the coherency in a greedy approach to construct reasonably coherent paragraphs, by finding the cut-off of a consecutive sequence of sentences. These resulting paragraphs starting and ending timestamps denotes the start and end of an aligned sequence of frames, and accordingly the generator yields a DataWindow holding an aligned sequence of frames with a segment of transcription.

\subsubsection{Pipeline}
Our pipeline starts with a `\textbf{KeyFrameExtractor}', that extracts representative key frames from the DataWindow aligned sequence of frames in a segment of time. We have adapted Kepler Lab Katna's \cite{katna_keplerlab} image selector \footnote{\url{https://katna.readthedocs.io/en/latest/_modules/Katna/image_selector.html}} method, with some modifications. The original algorithm expects a number of frames $n\_kframes$ to be picked, it applies few filtering mechanisms to filter out reasonable bad images in terms of brightness and contrast, then clusters those frame images into $n\_kframes$, given the histograms of gray-scaled version of each of those images to be clustered. Per each of those clusters, the key frames is identified accordingly by identifying the image that is associated with the highest image Laplacian variance score, pointing to it being least blurry member in the cluster.

Instead of passing $n\_kframes$, we take an automated approach to find a good number of clusters by adapting the scaled inertia approach suggested by Herman-Saffar  \cite{towardsdatascience_automated_kmeans_clusters} --- we also instead apply k-means using the FAISS library to speed up this relatively expensive process of picking the number of key frames \cite{johnson2019billion, douze2024faiss}. Furthermore, we omit the pre-filtering step by brightness and contrast for the aim of speed, given that also we do not need to worry about it as our number of clusters is dynamic, hence ideally these bad images will be avoided.


We run optical character recognition (OCR) on these representative key frames using `\textbf{EasyOCR}' \cite{easyocr_jaidedai}, implemented based on the work of \cite{shi2015endtoendtrainableneuralnetwork, baek2019characterregionawarenesstext}. Furthermore, ImageTagging runs as well on the same identified representative frames. Utilizing the RecognizeAnything model (`\textbf{RAM}') \cite{zhang2023recognizeanythingstrongimage}, we gather an understanding of all objects visible per each representative frame image. 

These recognized objects are only recognized as words at the point, which we use dynamically to construct a prompt using a simple `\textbf{ImgTextPromptDirector},' which acts as a PipeDirector for a `\textbf{GroundingDino}' \cite{liu2023grounding} pipe. The job of this pipe is to localize these objects, recognized by the ImageTagging pipe earlier, and possibly aided by OCR, by detecting the objects' corresponding bounding boxes.

Our goal at this point is to dense caption each of those representative frames images, supplemented by the knowledge we have about the present objects. We adapt part of the methodology applied by CaptionAnything \cite{wang2023caption} to gather as much captions as possible about the frame. We add a Pipe, `\textbf{HQEfficientSAM}' wrapping a high quality light variation of the SAM model \cite{sam_hq, transfiner} prompted with bounding boxes, generated by our previous Pipe, to get a fine-grained mask of each detected element. Those masks with their corresponding knowledge, accordingly are encapsulated in the DataWindow undergoing the processing, which gets passed down to the subsequent stages.

A `\textbf{CroppingObjectFocuser}' pipe utilizes the masks provided in the DataWindow at this stage. A number of ObjectFocusers have been tested qualitatively, and we did not see sufficient performance increase by cutting the image by exactly the mask (as by the approach performed in \cite{wang2023caption}), neither by having a black nor white background. Instead we crop the image by the rectangle bounding the given mask to branch out a frame image into smaller images each containing one or a subset of elements. As we do not cut by the mask, we can omit accordingly the segmentation step; however, it is not a bottleneck, and the boxes bounding the segmented masks offer finer bounding boxes around the elements of interest. Additionally, in some instances, the ImageTagging Pipe might fail to recognize any objects, and hence the grounding bounding boxes detection model is not prompted; however, the segmentation pipe is able to work around this drop in performance of previous pipes, by utilizing the automatic segment generation of the SAM model.

Moreover, those cropped images along with the complete image of the frame are extracted from the DataWindow by a director pipe, `\textbf{BranchingFocusedFramesDirector},' so that they can be captioned by a `\textbf{Captioner}' pipe employing the `blip-large base' \cite{li2022blipbootstrappinglanguageimagepretraining} model. We have assessed qualitatively the difference in performance between Blip and Blip2 for simple image captioning and we did not see in our experimental trials a boost in performance, in fact, we observed better performance with Blip in some close-up instances, while also it is significantly cheaper to run Blip. The captions are merged and repacked properly on the frame-level by concatenating them through `\textbf{MergingCaptionsPipe}.'

Finally, those dense captions generated are processed by a pipe, `\textbf{SentenceGraphParser}' that performs a sequence of textual processing techniques to extract the subjects, objects, and relationships identified in those captions. Those internal processing steps are as follows: 
\begin{enumerate}
    \item \textbf{Scene Graph Parsing} \cite{scene_graph_parser} A parser extracts the subjects, objects, and their associated relations, based on dependency parsing employing spacy roberta-based ``en\_core\_web\_trf'' language model \footnote{\url{https://huggingface.co/spacy/en_core_web_trf}}. Those relations' graphs are extracted and collected per every caption sentence in that dense caption paragraph per each frame in a DataWindow.

    \item \textbf{Co-Reference Resolution} Employing FastCoref \cite{Otmazgin2022FcorefFA}, we construct a dictionary of co-references on the scope of the frame (i.e., at each complete dense captions paragraph). Using the co-reference dictionary, we resolve the previously extracted scene graphs per each frame so that we can omit pronouns.

    \item \textbf{Concretness Filter} To filter the noisy relations in our graph, we adapt the approach performed in \cite{vaze2023genecisbenchmarkgeneralconditional}, by filtering the subjects-objects-relation tuples based on the mean of concreteness scores \cite{Brysbaert2014ConcretenessRF} of the subject, and the object. 
\end{enumerate}

\subsubsection{Consuming DataWindows: Videos KBs}

Our pipeline, if perceived as an end-to-end model, transforms a video into clauses, these clauses are made up of those subjects, objects, and relationships detected at this last stage. However, due to the design of our pipeline, and the structure of a framework, and the concept of a DataWindow, we consume these DataWindows at the end of the pipeline to construct a frame-level indexed knowledge base of the processed video. Consuming DataWindows this way, we transform a video into a semi-structured form of data including but not limited to detected objects and their relations, and time of appearance (associated with representative-frames in segments of the video). Hence, one can utilize these VideoKnowledgeBases to experiment, employ and analyze the collection/hierarchy of inferences and features extracted/generated by the employed models on our pipeline for downstream tasks, as demonstrated in the subsequent section.

\subsection{Query Objects from the Wild}\label{subsec:method_query_objects}
In this final phase, we convert VideoKnowledgeBases into query-able Video Knowledge Graphs, enabling retrieval of video segments based on multi-modal queries (text, image, etc.). By indexing nodes off the information we gathered from the collection of inferences throughout our pipeline recipe; and by connecting these nodes on the basis WordNet lexical relationships, we build a system that allows users to retrieve specific information in frames across videos. This approach supports adaptive, domain-specific extensions, enabling users to add new classifications. 

\subsubsection{Videos to Knowledge Graphs}
We employ our framework, and candidate pipeline based on the recipe from the previous section, we construct a database of VideoKnowledgeBases from a collection of videos of choice. Following that, we convert each of those VideoKnowledgeBases into what we call a VideoKnowledgeGraph, a knowledge graph consisting of nodes corresponding to Synsets with identifiers corresponding to indices of the associated frames along with accompanied inferences (i.e., bounding boxes, etc.) with respect to the DataWindow corresponding to the segment of the video where the particular knowledge is observed.

The key attribute that defines the nodes of the VideoKnowledgeBases is constructed off the transcription, detected objects tags, and generated dense captions per each representative-frame throughout each DataWindow from the constructed VideoKnowledgeBase. We extract nouns and verbs from those generated sentences and words. 

Moreover, with the help of PyWSD word sense disambiguation (WSD) algorithm \cite{pywsd14}, and relying on the context of the transcription across the whole video in addition to the captions and the tags on the DataWindow level, we are able to identify corresponding word senses on the basis of WordNet lexical english database \cite{wordnet_10.1145/219717.219748} --- accordingly for instance, from a node containing `car.n.01`, one can pin point to all segments from all videos containing a vehicle of type car in all videos.

The constructed SynsetNodes (Synset-based nodes) gets connected, by the $construct\_graph$ in algorithm \ref{alg:video_to_kg}, based on their WordNet hypernym/hyponym relations. For instance, $n_{policeman.n.01}$ gets connected to $n_{chef.n.01}$ given their lowest common hypernym $n_{person.n.01}$ from the WordNet lexicon database; If node $n_{person.n.01}$ is not present in the same graph, a new node is created, otherwise, the present node is retrieved. Accordingly we connect the three nodes' sub-graphs, and the indices are inherited appropriately (propagated from the children nodes to the parents). Furthermore, the frame-level graphs are merged, and subsequently the resulting DataWindow corresponding graphs are also merged to have a single graph for the processed video containing multi-indexed connected nodes.

\subsubsection{Frames Retrieval \& Information Appending}
\begin{figure}
    \centering
    \includegraphics[width=0.9\linewidth]{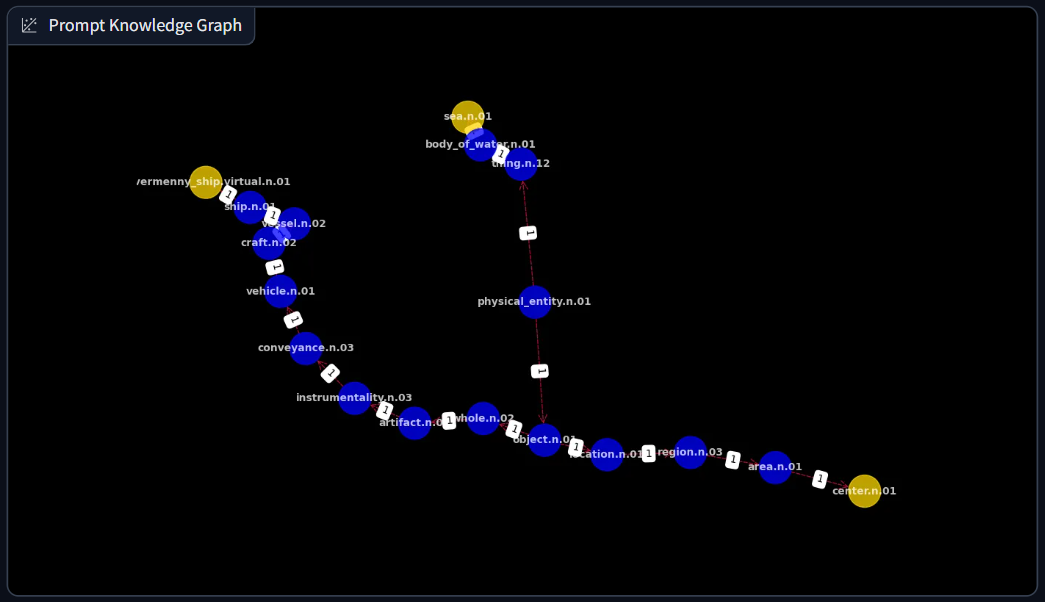}
    \caption{This figure illustrates a sample query, \textit{"a sovermenny ship in the middle of the sea"}, knowledge graph, representing the concept $ship$, its learned concept $sovermenny.ship.virtual.n.01$ and their relationships. The graph showcases the hierarchical structure that is used to query against the database of VideosKnoweldgeGraphs.
    }
    \label{fig:simple_query_graph}
\end{figure}

Applying the algorithm \ref{alg:video_to_kg}, discussed above, over a collection of videos, and building a simple interface to query the graphs, we are able to retrieve representative frames from any segments of all videos based on a multi-modal query (text, image, or a video). That is done by first converting this query into the same format (i.e., graph of the same hierarchy based on the same database lexicon as shown in the simple example in figure \ref{fig:simple_query_graph}), then we tests it against every VideoKnowledgeGraph in our database, based on how much is the query's graph overlapping the video's.

Furthermore, we define a `\textbf{VirtualSynset}`, which acts as a wrapper to new domain-specific-knowledge that can be appended to the knowledge graph (e.g., a user might want to append $kn95\_face\_mask.virtual.n.01$ to $face\_mask.n.01$, given that WordNet does not differentiate between the different models of face-mask, or does not include "kn95" particularly). Through our demo retrieval software, a user can create more specific novel words, and append them to the searchable graph database. 

We associate each of these new domain-specific nodes with a concept-level classifier. This is attainable by training a classifier, that is its only job is to differentiate between whether a face-mask is "kn95" or not. In our demo, qualitatively, we show that using about 50 samples, that are interactively annotated by the user in few seconds, our system fine-tunes a YOLOv8 \cite{yolov8_10533619} model to make this classification, which shall be hooked to our pipeline accordingly, so that any detected face-masks, gets further processed by this little classifiers, so that whenever a $face\_mask.n.01$ node gets constructed, we look up our VirtualSynsets database, and if possible extensions are available (i.e., virtual hyponym such as $kn95\_face\_mask.virtual.n.01$, or $surgical\_face\_mask.virtual.n.01$, we further test these masks under the associated VirtualSynsets' classifiers. Accordingly, once a new concept is added to the system, the system shall run in the background to update the existing graphs, employing the newly created fine-tuned mini-classifiers. A demo video is available in the supplementary materials.

\begin{algorithm*}
\caption{Converting VideoKnowledgeBase into VideoKnowledgeGraph}\label{alg:video_to_kg}
\begin{algorithmic}
\State $KB^{v} \gets Pipeline(v)$
\For{$kb_{i}$ in $KB^v$}  
    \For{$frame_{j}$ in $kb_{i}$}
        \State $words\_synsets\_dict \gets extract\_words(frame_{j})$
        \State $synsets\_per\_word \gets word\_sense\_disambiguation(words\_synsets\_dict, frame_{j}, kb_{i})$
        
        \State $nodes = []$
        \For{$s_{k}$ in $synsets\_per\_word$}
            \State $node \gets construct\_synset\_node(s_{k}, frame_{j})$ \Comment{$node$ is associated with relevant frame indices}
            \State add $node$ to the list of $nodes$ 
        \EndFor
        \State $g_{i_{k}} \gets construct\_graph(nodes)$
    \EndFor
    
    \State $g^{v}_{i} \gets merge\_graphs(g^{v}_{i_{k}})$, where $1 \leq k \leq |synsets\_per\_word|$
\EndFor
\State $G^{v} \gets merge\_graphs(g^{v}_{i})$, where $1 \leq i \leq |KB^{v}|$

\State \Return $G^{v}$
\end{algorithmic}
\end{algorithm*}
\section{Future Directions \& Applications}
In this work, we provide a novel approach for data retrieval across temporal multi-modal data, along a framework to speed-up the research, and ease down some engineering obstacles in analyzing videos, but there are still many other obstacles to tackle, and several avenues for future research and development.

\textbf{Key Observations.} In our experiments, we observed a few key areas for improvement. First, incorporating OCR can introduce noise into the pipeline, particularly when the OCR model misinterprets characters or struggles with complex visual scenes. This issue is exacerbated when dealing with animated text, where some frames may include only a partial version of the in-progress text, leading to inaccurate OCR results. Future work could focus on refining OCR integration by filtering out noisy results given context from other pipes in the pipeline.

Second, the dense captioning process, where we employ a single-image captioning model on a sequence of frames, can generate redundant captions. This redundancy arises because the model processes each frame independently, lacking the context to generate diverse captions, especially when consecutive frames have similar visual content. To address this, we propose providing the captioning model with context about previously generated captions. This could help the model generate more diverse and informative captions by considering the broader visual and textual context. 

\textbf{Untapped Potential.} The visual and auditory channels offer further opportunities for enhancing the framework. For instance, visual cues like scene changes and object movements could be incorporated into coherency calculation to better generate the DataWindows (i.e., segment the video) --- especially with silent video segments. Some work can be adapted from speaker diarization techniques \cite{10102534, sharma2022using} to better ground the transcription utterances and relevant captions, to bounded objects. Additionally, techniques from Visual Word Sense Disambiguation \cite{Kritharoula_2023} (VWSD) could leverage visual information to improve word sense disambiguation (e.g., utilizing ImageNet \cite{imagenet_cvpr09} with Clip \cite{wei2023iclip}; BabelNet \cite{navigli-ponzetto-2010-babelnet} already includes images representing senses).

\textbf{Embeddings and Knowledge Graph Enhancement.} We can explore translating discrete WordNet senses into continuous embedding space to enable more nuanced semantic comparisons \cite{9343669_continue_learning_embedding}. Another is to completely avoid discrete outcomes, and fuse an end to end scheme to utilize embeddings from the backbone of incorporated model rather decoding first to discrete vocabulary space. Furthermore, when calculating concreteness scores, we can use embedding similarity to estimate the score for missing words. Using BabelNet \cite{navigli-ponzetto-2010-babelnet}, with its multilingual and encyclopedic knowledge, could also enhance the Video Knowledge Graph. The frame-level structure of the graph can be further exploited to learn models that capture temporal context. 

\textbf{Learning the context across the DataWindows.} Given that we have sub-graphs on videos' frame-level, we can learn models that have an understanding of the context given a sequence of frame-level or data-window-level graphs (e.g., utilizing the clauses extracted by the Scene Graph Parser). Training an encoder model on the DataWindows generated by our pipeline holds significant potential. The encoder could learn to represent the dynamic relationships between entities and events over time, creating a comprehensive multi-modal representation of the video content. This could enable various downstream tasks, such as video summarization, event prediction, video retrieval, and question answering. 


\textbf{Applications and Use Cases.} The proposed pipeline can be used to generate datasets for training multimodal LLMs or specialized classifiers. It's important to address the challenge of potential noise in the generated data to ensure data quality.  The framework can also be adapted for use in Augmented Reality (AR) applications, particularly those involving socially intelligent agents \cite{lin2024estuary}. By incorporating real-time object detection and knowledge graph querying, a modified version of this pipeline could provide context to the agentic flow (e.g., augmenting the LLM context \cite{pi2023detgpt}), enabling more realistic and engaging interactions between users and virtual agents. For instance, virtual agents could react to recognized objects and events in the user's environment, provide relevant information or guidance, and engage in more meaningful conversations based on the user's context. This could lead to innovative AR experiences in various domains, such as education, entertainment, and social interaction. 
{
    \small
    \bibliographystyle{ieeenat_fullname}
    \bibliography{main}
}


\end{document}